\documentclass[letterpaper, 10 pt, conference]{ieeeconf}  
\IEEEoverridecommandlockouts                              
\usepackage{afterpage}

\usepackage{amsthm}
\usepackage{stmaryrd}
\usepackage{amsmath}
\usepackage{amssymb}
\usepackage{siunitx}
\usepackage{stfloats}
\usepackage{hyperref}
\hypersetup{
    colorlinks=true,
    linkcolor=blue,
    filecolor=magenta,      
    urlcolor=cyan,
}
\usepackage{graphicx}
\usepackage{gensymb}
\usepackage{caption}
\usepackage{subcaption}
\usepackage[table]{xcolor}
\usepackage{color}
\usepackage{svg}

\usepackage{multirow}
\usepackage{wrapfig,lipsum,booktabs}
\usepackage{makecell} 

\usepackage{booktabs} 
\usepackage{lipsum} 
\usepackage{pdfpages}
\usepackage{kotex}

\newcommand{\tss}[1]{{\tiny$\pm$#1}}
\title{\LARGE \bf
KineFuse: Kinematic-Aware Haptic Fusion for In-Hand Occluded-Object Pose Tracking

}

\author{Chanyoung Ahn, Jaesung Lee, Sungwoo Park, and Donghyun Hwang
\thanks{This work was supported in part by the Industrial Strategic Technology Development Program (RS-2024-00442029) funded by the Ministry of Trade, Industry and Energy (MOTIE, Korea), in part by the KIST Institutional Program, and in part by the Institute Strategic Development (ISD) Program. In Addition, this work was supported by the National Supercomputing Center with supercomputing resources including technical support (KSC-2025-CRE-0343). \textit{(Corresponding author: Donghyun Hwang)}}  \thanks{All authors are with the Center for Humanoid Research, KIST, Seoul, 02792 South Korea. Sungwoo Park and Jaesung Lee are also with Korea University, Seoul, 02841 South Korea. ({\tt\footnotesize \{chanyoung.ahn; jay.lee; sungwoo.park; donghyun\}@kist.re.kr}).}}

\begin{document}
\maketitle

\thispagestyle{empty}
\pagestyle{empty}
\begin{abstract}
Dexterous in-hand manipulation requires continuous 6D pose tracking, yet the manipulating fingers inevitably occlude the object from the camera. We study how to structure the sparse haptic signals already available on multi-fingered hands—proprioception, proximal force/torque, and binary contact—to complement a pretrained visual pose tracker under occlusion. We propose a kinematic-aware finger-level encoder and systematically compare it against four alternative designs through three levels of evaluation: per-frame refinement, sequential open-loop tracking, and closed-loop manipulation. Our experiments reveal that (i) per-frame evaluation cannot distinguish encoder quality, while sequential tracking amplifies architectural differences by up to 15 times; (ii) the structured encoder learns task-specific cross-modal gating—using vision exclusively for translation and dedicating one attention head to haptic for rotation—without explicit supervision; and (iii) compact finger-level tokenization (4 tokens) outperforms both flat fusion and joint-level representations, which suppress vision through norm dominance. We validate that improved tracking yields higher success in a downstream reorientation task and provide qualitative real-world demonstrations. Our project page is available at \url{https://cold-young.github.io/kine-fuse/}.



\end{abstract}
\section{Introduction}
\label{sec:intro}

Tasks such as soldering or handwriting require placing a tool tip at a precise target pose, demanding accurate real-time 6D tracking during in-hand reorientation~\cite{gordon1991integration, navarro2023visuo, bhardwaj2026viserdex}. Vision-based 6D pose estimation frameworks such as FoundationPose~\cite{wen2024foundationpose} and BundleSDF~\cite{wen2023bundlesdf} provide strong pose estimation backbones, but the in-hand regime exposes a fundamental weakness: the fingers that manipulate the object  also occlude it from the camera. Under self-occlusion, short-range depth artifacts, and constrained viewpoints, visual estimates degrade precisely when fine-grained control is most needed~\cite{mack2025visuo, rezazadeh2023hierarchical}. Prior systems address this through multi-camera rigs~\cite{andrychowicz2020learning, chen2023visual} or visibility-friendly task designs~\cite{wang2024penspin, qi2023general, yang2024anyrotate}, but such solutions restrict hardware or grasp repertoire.

\begin{figure}[t!]
\centering
\includegraphics[width=\linewidth]{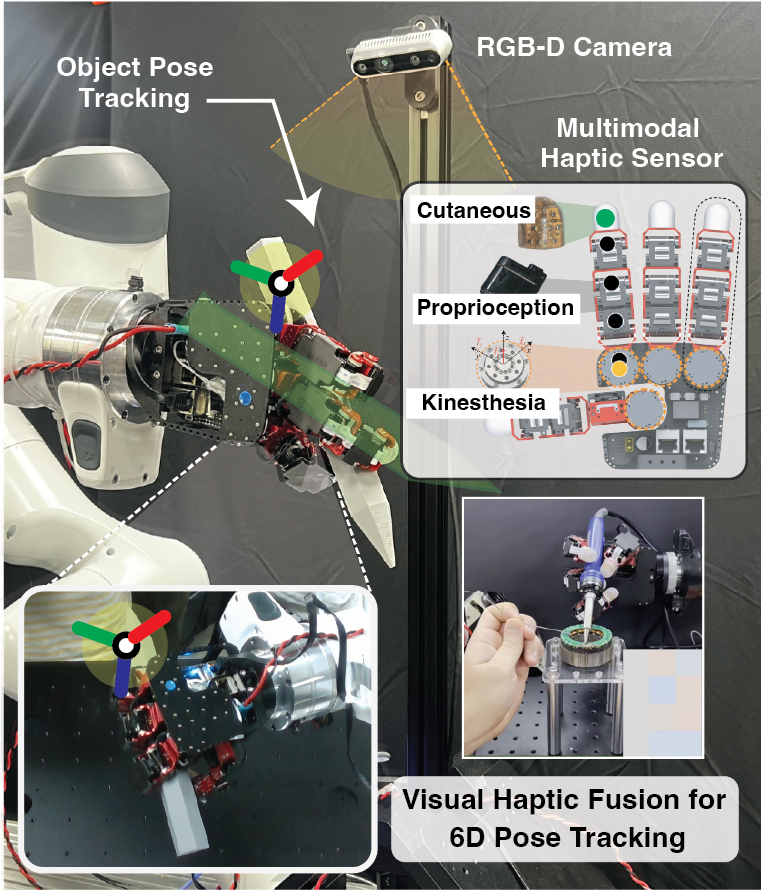}
\caption{Our KineFuse Framework combines vision with multimodal haptic feedback, such as tactile, force/torque and joint angle, to enable robust 6D pose tracking during dexterous in-hand manipulation under occlusion.}
\label{fig:concept}
\vspace{-1.em}
\end{figure} 


A growing body of work therefore combines visual and tactile feedback for in-hand 6D pose estimation~\cite{rezazadeh2023hierarchical, wang2024penspin, tu2023posefusion, wanvint, suresh2024neuralfeels}. While effective, many such methods rely on dense vision-based tactile sensors such as GelSight~\cite{yuan2017gelsight} or DIGIT~\cite{lambeta2020digit}, which provide rich contact images but require specialized hardware. By contrast, many multi-fingered hands already provide simpler embodied haptic signals: joint-level proprioception, binary contact, and force/torque (F/T) measurements. These signals are far more accessible, but when used at all they are often fused as a flat vector~\cite{lin2025learning}, discarding the kinematic and spatial relationships among sensing sites. Recent work further suggests that even minimal contact signals can improve pose estimation under occlusion when incorporated appropriately~\cite{mack2025visuo}. This points to a central question: not merely whether haptics should be added, but how sparse haptic signals should be structured.

In this work, we study how structured encoding of sparse  embodied haptics—proprioception, proximal F/T, and binary contact—can complement a pretrained visual tracker under occlusion. Our key idea is to preserve the hand's kinematic structure when encoding haptic observations, rather than treating them as an unstructured vector (Fig.~\ref{fig:concept}). We keep FoundationPose~\cite{wen2024foundationpose} frozen and augment it with a URDF-aware finger-level haptic encoder. Through a systematic comparison of five encoder designs across three evaluation levels—per-frame, sequential tracking, and downstream manipulation—we identify when and how structured  haptic encoding improves occlusion robustness.

In summary, our contributions are:
\begin{enumerate}
\item \textbf{Kinematic-aware haptic encoder and ablation.}
We propose a URDF-aware finger-level encoder for sparse 
haptic signals and systematically compare five encoder 
designs, identifying that compact 4-token finger-level 
representations with URDF spatial bias outperform both 
flat fusion and joint-level encodings.

\item \textbf{Multi-level evaluation revealing gap amplification.}
We evaluate across per-frame, sequential tracking, and 
downstream manipulation, showing that architectural 
differences invisible at the per-frame level (1.12 times) 
amplify to 2 times in tracking and 15 times in 
manipulation—establishing that sequential evaluation is 
necessary for meaningful assessment of fusion architectures.

\item \textbf{Analysis of learned fusion behavior.}
We analyze the learned attention patterns and find that 
the structured encoder learns physically interpretable 
cross-modal gating: translation relies on vision exclusively 
while rotation dedicates a specialized attention head to 
haptic tokens.
\end{enumerate}
\section{Related Work}
\label{sec:rel}

\subsection{In-Hand Pose Tracking}

Vision-based 6D object pose estimation and tracking have advanced rapidly, with frameworks such as BundleSDF, FoundationPose, and SAM-6D serving as strong perception backbones for robotic manipulation~\cite{wen2024foundationpose, wen2023bundlesdf, lin2024sam}. Building on these advances, several works have addressed in-hand reorientation and dexterous manipulation using vision-based pose tracking~\cite{rezazadeh2023hierarchical, wang2024penspin, tu2023posefusion, wanvint, suresh2024neuralfeels}.

However, the in-hand regime remains especially challenging for vision-only perception. Self-occlusion by the fingers and constrained wrist-mounted viewpoints often degrade pose estimates during contact-rich manipulation~\cite{chen2023visual}. Prior systems mitigate this with multi-camera setups~\cite{andrychowicz2020learning, chen2021system}, finger-gaiting strategies~\cite{chen2023visual, wang2024penspin, yang2024anyrotate, pmlr-v205-qi23a} that periodically expose the object, or task designs that improve visibility~\cite{wang2024penspin, pmlr-v205-qi23a, yang2024anyrotate}, but such solutions can restrict the grasp repertoire or hardware configuration.

A growing body of work therefore combines vision with tactile or force-related feedback for dexterous manipulation and in-hand pose tracking~\cite{mack2025visuo, rezazadeh2023hierarchical, li2025v, dikhale2022visuotactile, oller2023manipulation, hu2025dexterous}. Recent optimization-based work~\cite{mack2025visuo}, for example, incorporates binary contact as a geometric constraint in a factor-graph optimizer, showing that even minimal tactile information can improve pose estimation under occlusion.

Despite this progress, two limitations remain. First, many methods rely on vision-based tactile sensors such as GelSight~\cite{yuan2017gelsight} and DIGIT~\cite{lambeta2020digit}, leaving lower-dimensional signals, including joint-level proprioception, binary contact, and, when available, proximal force/torque, comparatively underexplored for structured fusion. Second, with some notable exceptions~\cite{rezazadeh2023hierarchical}, heterogeneous sensor streams are often fused through straightforward concatenation, which does not explicitly encode the kinematic or spatial relationships among sensing sites on the hand. 

\subsection{Structured Haptic Representations}

Beyond sensor choice, the effectiveness of multimodal in-hand perception also depends on how haptic signals are represented and fused. Existing approaches span several design choices. The simplest approaches concatenate raw sensor readings into a flat vector for a Multi-Layer Perceptron (MLP)~\cite{lin2025learning}, which is lightweight but discards spatial and kinematic relationships among sensing sites. Richer approaches embed hand--object interaction into shared 3D representations such as point clouds with annotated contact regions~\cite{suresh2024neuralfeels, li2025v, dikhale2022visuotactile, watkins2019multi,  falco2017cross}, providing stronger geometric bias at the cost of additional rendering or reconstruction pipelines. A third line of work represents the hand as a kinematic graph and processes it with GNNs~\cite{rezazadeh2023hierarchical, yang2023tacgnn, rampavsek2022recipe} or graph transformers~\cite{ying2021transformers, patel2025get}, leveraging the fact that sensors are physically attached to specific links and joints.

Among graph-based approaches, Rezazadeh et al.~\cite{rezazadeh2023hierarchical} are the closest precedent: they construct a heterogeneous graph over vision, tactile, and proprioceptive nodes for in-hand object pose estimation. However, like most visuo-haptic frameworks, their method relies on dense vision-based tactile sensing. Recent optimization-based work by Mack et al.~\cite{mack2025visuo} shows that low-resolution tactile contact can also improve in-hand pose estimation under occlusion, but does so through factor-graph-based geometric correction rather than learned multimodal fusion. Whether graph-structured encoding remains effective in the sparse regime—where only a subset of joints carry force/torque measurements and the remaining joints provide only proprioception—remains unclear. 

To our knowledge, prior work has not systematically analyzed how the value of structured haptic fusion changes across graded occlusion severity relative to both vision-only and naive fusion baselines. These gaps motivate our focus on structured sparse-haptic encoding for occlusion-robust in-hand pose tracking.

\section{Method}\label{sec:method}

\begin{figure*}[t!]
\center
\includegraphics[width=0.85\textwidth]{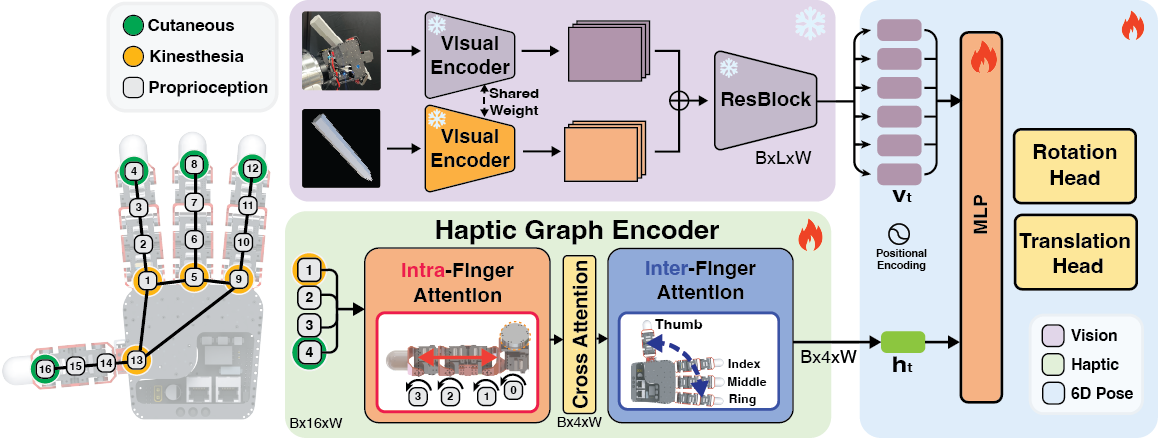}
\caption{\textbf{KineFuse framework.} A shared visual encoder produces $L{=}400$ tokens from rendered and observed RGB-D crops; the Finger Graph Encoder maps 16-joint haptic signals to $K{=}4$ finger tokens. Both are concatenated and decoded by separate translation and rotation heads.}
\label{fig:2-overall}
\end{figure*}

\begin{figure}[t!]
    \center
    \includegraphics[width=0.9\linewidth]{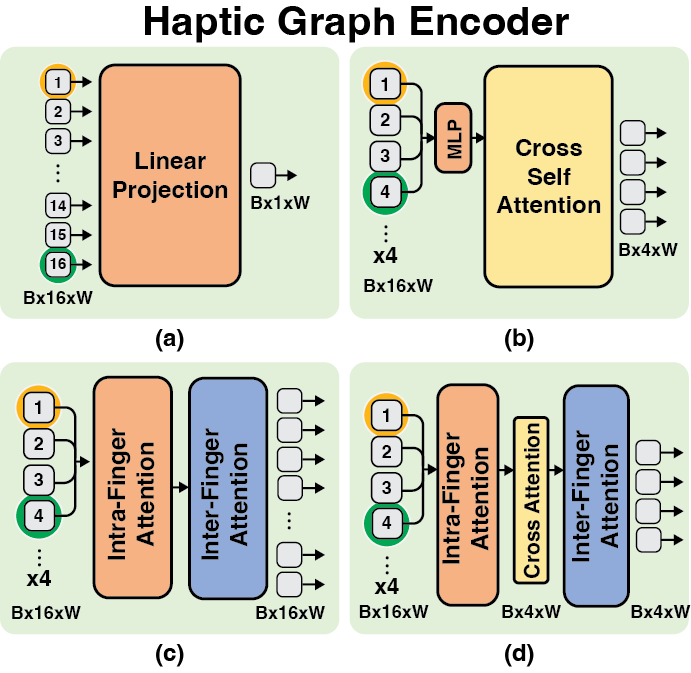}
    \caption{\textbf{Haptic Graph Encoder.} (a)~Naive, (b)~FingerMLP, (c)~16-token, and (d)~KineFuse: intra-finger attention with a block-diagonal mask propagates sparse F/T cues within each finger, cross-attention pools joints into finger tokens, and Graphormer layers with URDF spatial bias capture inter-finger relationships.}
    \label{fig-2-1:ablation}
\vspace{-1.0em}
\end{figure}

\subsection{Problem Formulation}\label{sec:formulation}

We consider 6D object pose tracking during in-hand manipulation with an external RGB-D camera and no external tracking system (see in Fig.~\ref{fig:2-overall}). At each timestep~$t$, the system receives an RGB-D observation~$I_t$ and a sparse haptic observation~$h_t$, and predicts the object pose $\hat{T}_t \in SE(3)$ in the camera frame. The haptic input is collected from a $J$-joint dexterous hand over $\tau$ history steps. We represent it as $h_t \in \mathbb{R}^{J \times \tau \times C}$, where $J$ is the number of joints and $C$ is the per-joint feature dimension. 

In our setup, $J{=}16$ and $\tau{=}4$; the short temporal window captures short-term variations in force and contact signals during in-hand interaction. All joints provide proprioceptive states, namely joint position~$q$ and velocity~$\dot{q}$. In addition, binary tactile contact is measured at the fingertips, while a subset of phalangeal joints provides 3-axis proximal force/torque~(F/T) measurements together with a binary availability flag that indicates whether F/T sensing is present at that joint.

Let $\mathcal{S}_{\mathrm{FT}} \subset \{1,\dots,J\}$ denote the joints equipped with F/T sensing. In our hand, only four phalangeal joints belong to $\mathcal{S}_{\mathrm{FT}}$, while the remaining joints provide proprioception only. 
This heterogeneous sensing layout---combining joint-level proprioception, fingertip contact sensing, and sparse force measurements at only a subset of joints---defines the sparse haptic regime that motivates our structured encoding approach. 
We assume a known object CAD model and known hand--eye calibration~\cite{marchand2006visp} between the camera and robot base frames. 

\subsection{Visual Pose Estimation}\label{sec:visual}

We build on the FoundationPose refinement network~\cite{wen2024foundationpose} as the visual backbone (see in Fig.~\ref{fig:2-overall}). Given a current pose hypothesis $\tilde{T}_t$, the object mesh is rendered at $\tilde{T}_t$ and compared against the observed RGB-D input $I_t$ in a render-and-compare pipeline. Following FoundationPose, we crop both the rendered and observed inputs around the projected object region and encode them with a shared CNN backbone, producing a sequence of visual tokens $\mathbf{v}_t \in \mathbb{R}^{L \times D}$ where $L=400$ (a $20{\times}20$ spatial grid) and $D=512$ in our implementation. 
\subsection{Structured Haptic Encoding}\label{sec:haptic}
We compare five haptic encoder designs that progressively 
introduce kinematic structure (Fig.~\ref{fig-2-1:ablation}): 
(a)~\textbf{Naive} concatenates all joint features into a single token via linear projection; (b)~\textbf{FingerMLP} groups joints by finger and applies per-finger MLPs followed by cross-finger self-attention, producing 4 tokens without kinematic bias; (c)~\textbf{16-token} applies intra-finger and inter-finger attention at the joint level without pooling, retaining all 16 tokens; (d)~\textbf{KineFuse (Ours)} adds intra-finger attention, learned cross-attention pooling (16$\to$4), and URDF-aware inter-finger graph attention.  All variants share the same joint tokenization and fusion architecture  (Section~\ref{sec:fusion}); only the haptic encoding differs. 
We detail our full encoder below.

\textbf{Joint tokenization.} For each joint $j$, we flatten its $\tau$-step history into a feature vector $x_t^{(j)} \in \mathbb{R}^{28}$ and map it into the token space through a shared two-layer MLP with layer normalization. $\hat{z}_t^{(j)} = W_2 \,\phi\!\left(\mathrm{LN}(W_1 x_t^{(j)})\right)$ where $W_1: \mathbb{R}^{28}\!\to\!\mathbb{R}^{512}$, $W_2: \mathbb{R}^{512}\!\to\!\mathbb{R}^{512}$, $\mathrm{LN}$ denotes LayerNorm, and $\phi$ is GELU. We then add a learnable finger-identity embedding and a learnable within-finger position embedding. $z_t^{(j)} = \hat{z}_t^{(j)} + e_{\mathrm{finger}}(j) + e_{\mathrm{pos}}(j)$. This produces $J$ joint-level tokens $\mathbf{z}_t^{(0)} \in \mathbb{R}^{J \times 512}$.

\textbf{Intra-finger attention.} To propagate sparse force and contact cues within each finger, we apply two self-attention layers with a finger-restricted mask. $ \mathbf{z}_t^{(1)} = \mathrm{IntraAttn}(\mathbf{z}_t^{(0)}; M_{\mathrm{finger}})$
where $M_{\mathrm{finger}}$ blocks cross-finger attention. Each joint can attend only to the four joints of its own finger, allowing force and contact information available at distal sensing sites to propagate to proximal joints that lack direct force sensing.

\textbf{Finger-level pooling.} We compress the 16 joint tokens into 4 finger tokens using learned cross-attention. For each finger, a learnable query attends over that finger's four joint tokens and produces a single summary token. $\mathbf{f}_t = \mathrm{FingerPool}(\mathbf{z}_t^{(1)}) \in \mathbb{R}^{4 \times 512}$.

\textbf{Inter-finger graph attention.} The finger tokens are then processed by two Graphormer-style layers~\cite{ying2021transformers} augmented with URDF-derived spatial biases, following the embodiment-aware bias design of GET-Zero~\cite{patel2025get} $\mathbf{h}_t = \mathrm{InterAttn}(\mathbf{f}_t; B)$.
where $B \in \mathbb{R}^{4 \times 4 \times H}$ is a learned multi-head attention bias. For each finger pair, we compute a small set of hand-geometry features from the URDF, including hop distance, opposition relationship, and adjacency, and map them through an MLP to per-head bias terms. The final haptic representation is $\mathbf{h}_t \in \mathbb{R}^{4 \times 512}$.

\subsection{Visuo-Haptic Fusion}\label{sec:fusion}

We fuse the two modalities by concatenating the visual and haptic tokens along the sequence dimension:
\begin{equation}
\mathbf{z}_t = [\,\mathbf{v}_t;\mathbf{h}_t\,] \in \mathbb{R}^{(L+4)\times 512}.
\end{equation}
The fused token sequence is processed by two single-layer transformer encoders, one for translation and one for rotation:
\begin{align}
\mathbf{z}_t^{\mathrm{trans}} &= \mathrm{TransHead}(\mathbf{z}_t),\\
\mathbf{z}_t^{\mathrm{rot}}   &= \mathrm{RotHead}(\mathbf{z}_t).
\end{align}
We then mean-pool each sequence and project to the output space:
\begin{align}
\Delta \hat{\mathbf{t}}_t &= \tanh\!\bigl(W_{\mathrm{trans}}\, \mathrm{mean}(\mathbf{z}_t^{\mathrm{trans}})\bigr) \odot \mathbf{n}_t,\\
\Delta \hat{\mathbf{r}}_t &= W_{\mathrm{rot}}\, \mathrm{mean}(\mathbf{z}_t^{\mathrm{rot}}),
\end{align}
where $\Delta \hat{\mathbf{t}}_t \in \mathbb{R}^{3}$ is the translation update, $\mathbf{n}_t$ is a per-axis normalizer that bounds the maximum translation step size, and $\Delta \hat{\mathbf{r}}_t \in \mathbb{R}^{6}$ is the rotation update in the 6D representation~\cite{zhou2019continuity}. Both updates are predicted in the egocentric frame of the current hypothesis. The refined pose is obtained by composition: $\hat{T}_t = \Delta \hat{T}_t \circ \tilde{T}_t$. 

At inference we apply two successive refinement iterations: the output of the first pass serves as $\tilde{T}_t$ for a second pass, allowing the model to correct residual errors. During tracking, the initial hypothesis is set to the previous frame's estimate, $\tilde{T}_t = \hat{T}_{t-1}$.

Although the codebase supports a gated dual-head variant with a haptic-only prediction branch, we found that the confidence gate consistently saturated toward the haptic-only head, blocking visual gradients and collapsing to unimodal prediction. All reported results therefore use the direct fusion output without gating.

\subsection{Training}\label{sec:training}

\textbf{Two-stage training.} We train the model in two stages. In Stage~1, we pretrain the haptic encoder with a haptic-only pose prediction objective, ensuring that the haptic tokens carry a meaningful sparse-haptic representation before fusion. In Stage~2, we combine pretrained FoundationPose vision weights with the Stage~1 haptic encoder and fine-tune the full visuo-haptic model. During the first three warmup epochs of Stage~2, the visual backbone is frozen; afterward, all parameters are optimized jointly. The haptic branch uses a $2.5$ times learning rate relative to the visual backbone.

\textbf{Training pairs.} Following the FoundationPose refinement protocol, we perturb the ground-truth pose $T_t^*$ with sampled egocentric noise---rotation uniformly drawn from $[1^\circ, 10^\circ]$ and translation from $[0.002, 0.01]$\,m---to obtain a pose hypothesis $\tilde{T}_t$. The rendered crop is generated from $\tilde{T}_t$, while the observed RGB-D image and haptic signals correspond to the actual scene. The model predicts the residual pose delta that refines $\tilde{T}_t$ toward $T_t^*$.

\textbf{Data and augmentation.} Training data are collected in IsaacLab~\cite{mittal2025isaaclab} from in-hand reorientation trajectories under both clean and physically occluded settings. We apply synthetic rectangular occlusion to the observed RGB-D input and domain randomization to the force channels, including multiplicative scaling, additive offset, Gaussian noise, and stochastic dropout.

\textbf{Losses.} The total training objective is
\begin{equation}
\mathcal{L}
=
\mathcal{L}_{\mathrm{pose}}
+\lambda_{\mathrm{ADD}}\,\mathcal{L}_{\mathrm{ADD}}
+\lambda_{\mathrm{attr}}\,\mathcal{L}_{\mathrm{attr}}
+\lambda_{\mathrm{pen}}\,\mathcal{L}_{\mathrm{pen}}.
\end{equation}


Here, $\mathcal{L}_{\mathrm{pose}}$ denotes the MSE loss on the predicted egocentric translation and rotation deltas. The term $\mathcal{L}_{\mathrm{ADD}}$ is the standard Average Distance of Distinguishable model points (ADD) loss computed on the transformed object points. The remaining two terms provide auxiliary geometric supervision: $\mathcal{L}_{\mathrm{attr}}$ encourages hand--object proximity around the intended contact regions, while $\mathcal{L}_{\mathrm{pen}}$ penalizes mesh interpenetration. These auxiliary losses follow prior object--hand geometric supervision~\cite{li2025v}. The coefficients $\lambda_{\mathrm{ADD}}$, $\lambda_{\mathrm{attr}}$, and $\lambda_{\mathrm{pen}}$ balance the relative contributions of the corresponding loss terms.

\section{Experiments}\label{sec:experiments}

We design our experiments to answer four questions:
\begin{enumerate}
\item How does structured sparse haptic fusion compare to vision-only and naive fusion across occlusion severity?
\item Does improved pose tracking remain robust under manipulation-relevant trajectories?
\item Which aspects of the haptic encoder matter most---its structure, token count, or both?
\item Which haptic modalities contribute most---proprioception, F/T, or contact?
\end{enumerate}

\subsection{Experimental Setup}\label{sec:setup}

\begin{figure}[t!]
    \includegraphics[width=0.9\linewidth]{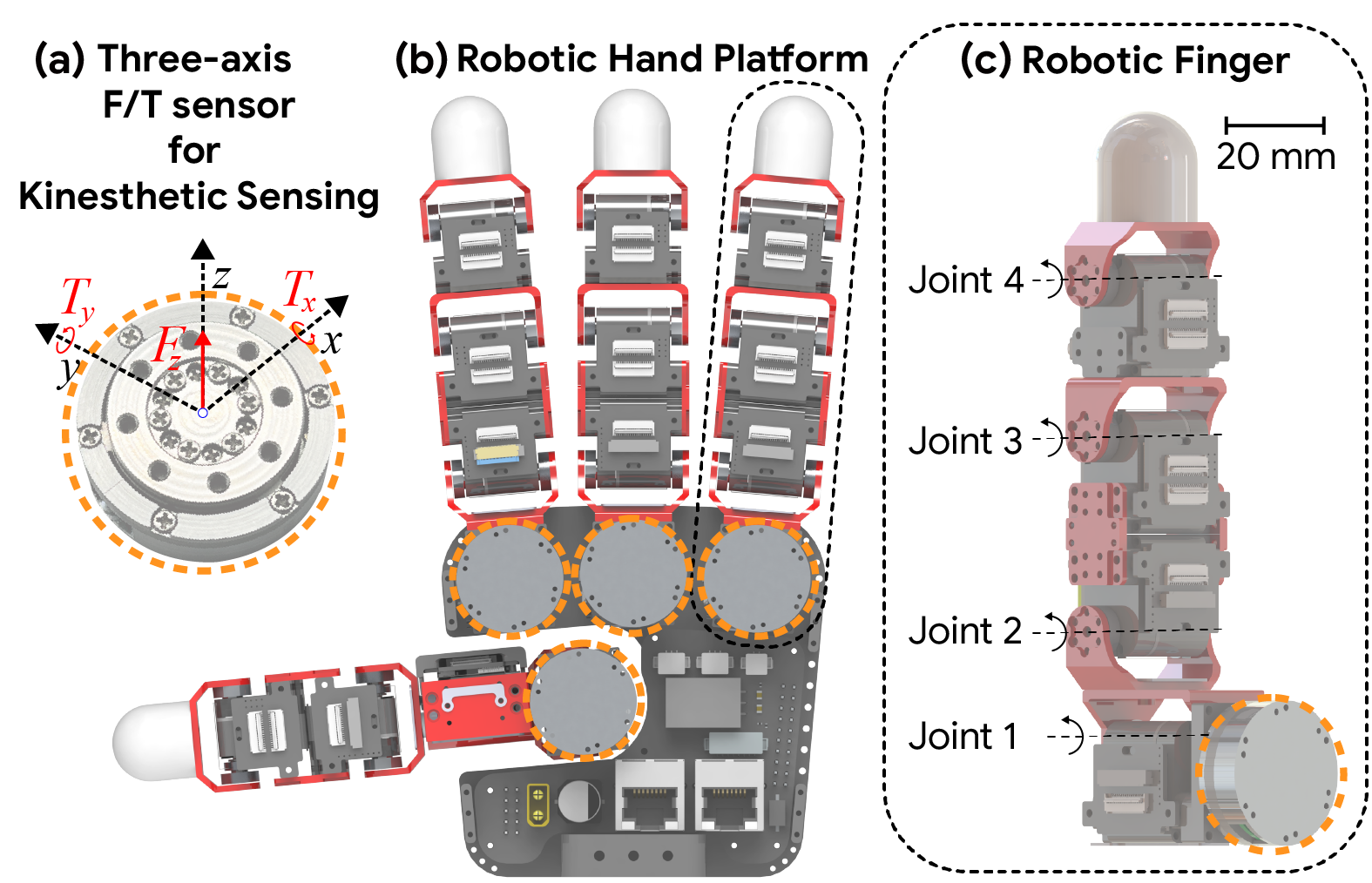}
    \caption{Illustration of robotic hand platform capable of kinesthetic feedback at each finger. (a), (b), and (c) show the three-axis F/T sensor embedded into the robotic hand platform for kinesthetic feedback, the robotic hand platform, and the robotic finger that includes tactile sensors respectively.}
    \label{fig-4:kistar}
\vspace{-0.5em}
\end{figure}

\begin{figure}[t]
\centering
\includegraphics[width=\linewidth]{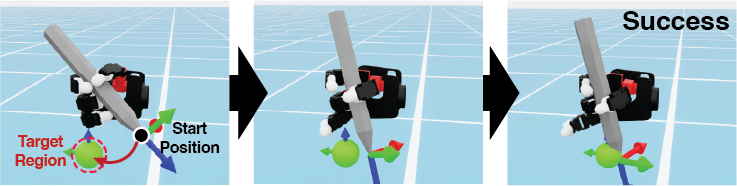}
\caption{{\textbf{Tool Reorientation Task.} The policy reorients a grasped tool to align its tip with a target point. We evaluate whether improved pose tracking under occlusion translates to higher task success.}}
\label{fig:rl_tasks}
\end{figure}

\begin{figure}[t!]
    \includegraphics[width=\linewidth]{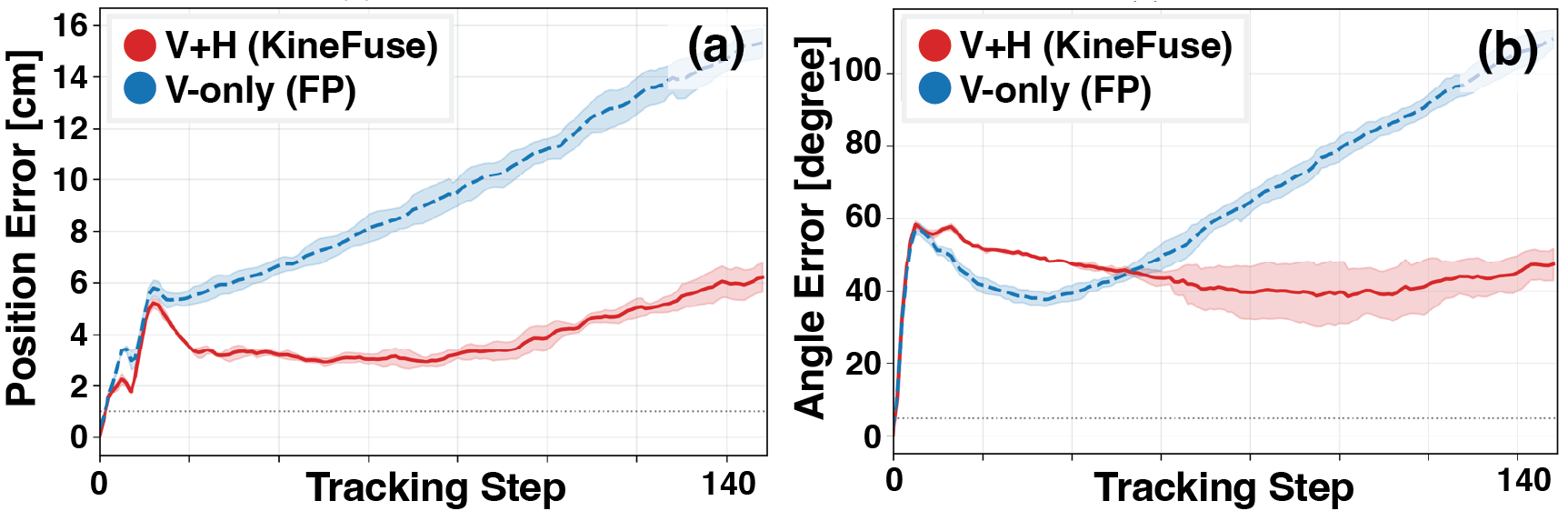}
    \caption{\textbf{Tracking drift.} Position and angular error over 149 tracking steps at 30\% occlusion. KineFuse (V+H) stabilizes after step 20, while V-only (FP) diverges 
continuously.}
    \label{fig:tracking-err}
\vspace{-0.5em}
\end{figure}

\begin{figure}[t!]
    \center
    \includegraphics[width=0.85\linewidth]{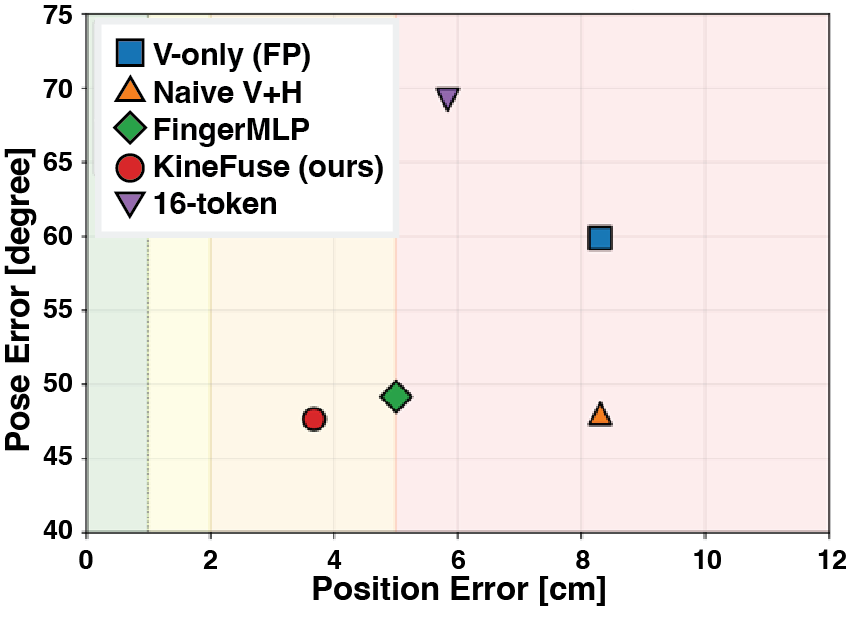}
    \caption{\textbf{Model errors in pose-tolerance space.} Each model's mean tracking error is overlaid on the RL policy's noise-tolerance curve. KineFuse operates within  the viable region; V-only and 16-token fall outside it.}
    \label{fig:tolerrence_eval}
\vspace{-0.5em}
\end{figure}

\begin{figure*}[t!]
\center
\includegraphics[width=0.9\textwidth]{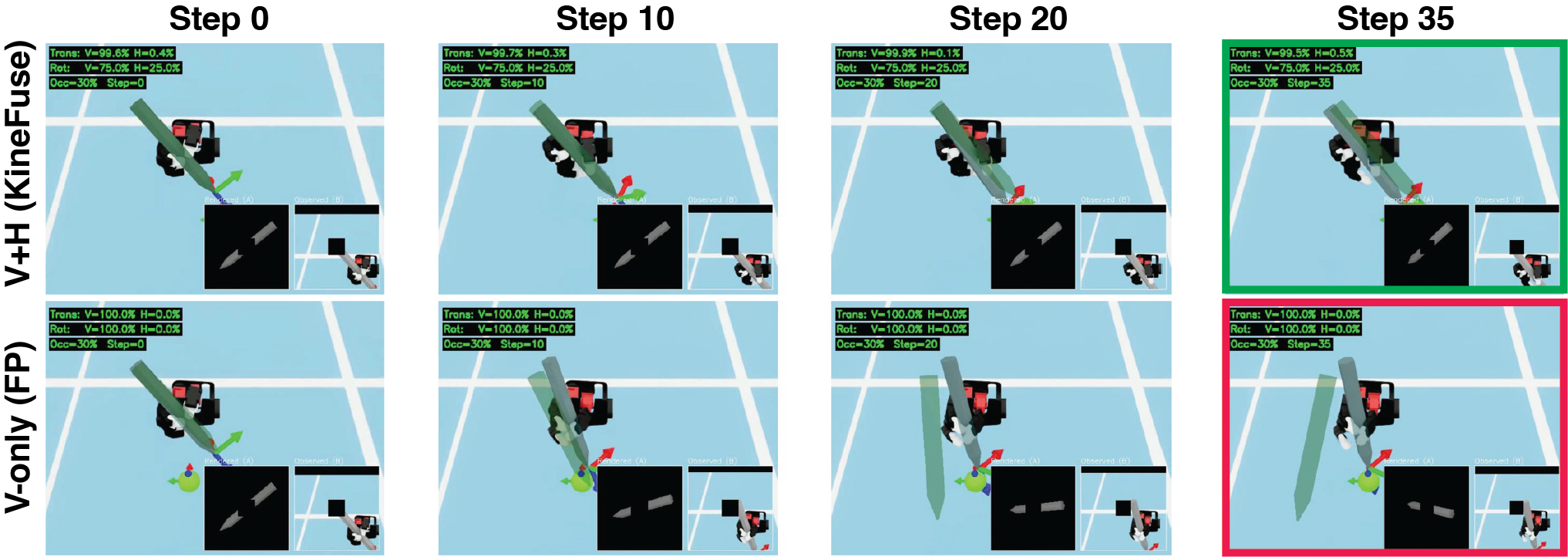}
\caption{\textbf{Qualitative tracking comparison under 30\% 
occlusion.} Top: KineFuse (V+H) maintains accurate pose 
overlay through step 35. Bottom: V-only (FP) diverges by 
step 20 as the occluded region grows. Insets show the 
rendered hypothesis (A) and observed crop (B) fed to the 
refiner. Attention allocation percentages (green text) 
confirm that KineFuse uses 100\% vision for translation 
and 75/25\% vision/haptic for rotation throughout.}
\label{fig:qualitative}
\end{figure*}


\textbf{Implementation details.} All models, including the V-only FoundationPose baseline, are fine-tuned with identical training schedules, datasets, and augmentation pipelines; only the haptic encoder differs. We report position error in centimeters (cm), angular error in degrees, and ADD in centimeters. Task success is measured as the mean count of successful tip-target alignments (within 2 cm position and 15 degrees orientation) per episode of 300 steps. The per-axis translation normalizer is set to [0.01, 0.01, 0.01]m, bounding the maximum single-step translation update.

\textbf{Hardware.} We use a four-fingered dexterous hand platform~\cite{park2024three} with 16 degrees of freedom (4 joints per finger) in detailed Fig.~\ref{fig-4:kistar}. Each joint provides encoder-based position and velocity readings. The four fingertip joints additionally carry 3-axis proximal force/torque sensors and binary contact indicators. An RGB-D camera (RealSense D435i for real-world) is mounted at the wrist. The manipulated object is a pencil shape with a known CAD model. 

\begin{figure}[t]
\centering
\includegraphics[width=0.9\linewidth]{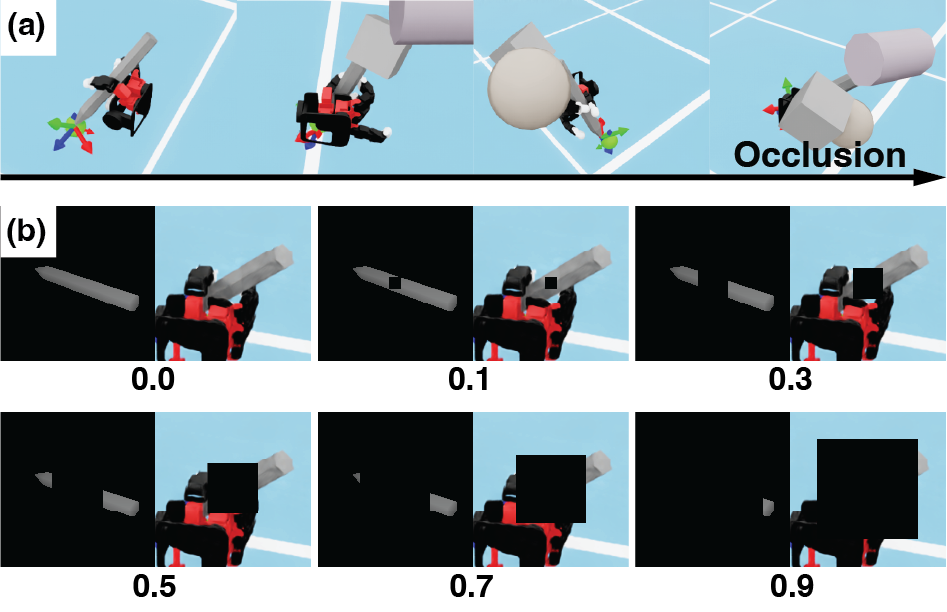}
\caption{\textbf{Occlusion Level Settings.} 
(a)~Physical occlusion in simulation: objects of increasing size 
are placed between the camera and the manipulated object. 
(b)~Synthetic rectangular occlusion applied to the observed 
RGB-D crop at varying ratios (0.0--0.9). Both settings are 
used during training; evaluation sweeps the occlusion ratio 
to measure robustness under progressive visual degradation.}
\label{fig:occ_settings}
\vspace{-1.0em}
\end{figure}

\textbf{Simulation.} Our manipulation goal is that reorients a grasped tool to align its tip to target region (see Fig.~\ref{fig:rl_tasks}). All training and quantitative evaluation are conducted in IsaacLab (Isaac Sim) \cite{mittal2025isaaclab}. We collect two datasets: \emph{clean} sequences (${\sim}$2{,}000 frames) of RL-driven in-hand reorientation without deliberate occlusion, and \emph{occluded} sequences (${\sim}$1{,}000 frames) with physical occluder primitives inserted between the camera and object. Ground-truth 6D poses are available for every frame (Fig.~\ref{fig:occ_settings} (a)).

\textbf{Occlusion protocol.} To evaluate robustness across graded occlusion severity, we apply synthetic rectangular masks to the RGB-D input at evaluation time, occluding approximately $0\%$, $10\%$, $30\%$, $50\%$, $70\%$,  and $90\%$ of the image area centered around the object region (Fig.~\ref{fig:occ_settings} (b)). This provides controlled, reproducible occlusion conditions independent of the physical occluder placement used during training (Fig.~\ref{fig:occ_settings}).

\textbf{Baselines and models.} We compare the following models in Fig~\ref{fig-2-1:ablation}, all built on the same frozen FoundationPose visual backbone and trained on the same combined dataset with identical augmentation. 

\textbf{Metrics.} For pose estimation, we report position error~(cm), angular error~(degree), and ADD~(cm), the average distance between object model points transformed by the predicted and ground-truth poses. For downstream manipulation, we report task success rate and mean successes per episode from a separately trained RL reorientation policy evaluated with estimated (rather than ground-truth) pose observations.

\begin{table}[t]
\caption{Open-loop sequential tracking under occlusion 
(3 seeds $\pm$ std). Pose at frame $t$ initializes frame $t{+}1$, so errors accumulate. KineFuse achieves the lowest position error at every occlusion level and  maintains constant angular error (47.6 degrees). KineFuse is the only model that maintains constant angular error (47.6 degrees) regardless of occlusion level.}
\label{tab:openloop}
\centering
\setlength{\tabcolsep}{2.5pt}
\footnotesize
\begin{tabular}{ll|cccccc}
\toprule
& & \multicolumn{6}{c}{Occlusion Ratio} \\
Model & Metric & 0\% & 10\% & 30\% & 50\% & 70\% & 90\% \\
\midrule
V-only 
  & Pos & 8.30\tss{.51} & 6.80\tss{.25} & 9.12\tss{.54} 
        & 7.59\tss{.45} & 9.68\tss{.73} & 7.14\tss{.44} \\
  & Ang & 59.9\tss{3.2} & 64.2\tss{3.4} & 57.7\tss{3.4} 
        & 60.6\tss{3.5} & 54.9\tss{1.3} & 79.4\tss{3.7} \\
\midrule
Naive 
  & Pos & 8.30\tss{.04} & 7.76\tss{.18} & 9.13\tss{.21} 
        & 10.1\tss{.17} & 7.01\tss{.20} & 5.57\tss{.19} \\
  & Ang & 48.0\tss{1.1} & 47.3\tss{1.1} & 47.9\tss{1.5} 
        & 47.7\tss{1.3} & 50.8\tss{1.2} & 53.7\tss{1.4} \\
\midrule
FingerMLP 
  & Pos & 5.01\tss{.23} & 5.46\tss{.32} & 5.35\tss{.28} 
        & 6.07\tss{.34} & 4.84\tss{.24} & 6.50\tss{.62} \\
  & Ang & 49.2\tss{4.2} & 52.8\tss{5.1} & 49.7\tss{3.8} 
        & 50.9\tss{4.3} & 47.0\tss{1.8} & 67.6\tss{4.2} \\
\midrule
16-token 
  & Pos & 5.84\tss{.08} & 5.84\tss{.08} & 5.81\tss{.08} 
        & 5.79\tss{.08} & 5.80\tss{.08} & 5.79\tss{.08} \\
  & Ang & 69.2\tss{1.3} & 69.3\tss{1.3} & 69.3\tss{1.3} 
        & 69.3\tss{1.3} & 69.3\tss{1.3} & 69.2\tss{1.3} \\
\midrule
\textbf{KineFuse} 
  & Pos & \textbf{3.68}\tss{.14} & \textbf{3.49}\tss{.10} 
        & \textbf{3.47}\tss{.12} & \textbf{4.17}\tss{.06} 
        & \textbf{4.33}\tss{.06} & \textbf{5.11}\tss{.07} \\
  & Ang & \textbf{47.7}\tss{2.7} & \textbf{47.7}\tss{2.7} 
        & \textbf{47.6}\tss{2.7} & \textbf{47.7}\tss{2.7} 
        & \textbf{47.6}\tss{2.7} & \textbf{47.6}\tss{2.7} \\
\bottomrule
\end{tabular}
\vspace{-1em}
\end{table}

\begin{table}[t]
\caption{Downstream manipulation success (success/episode, RL policy, 3 seeds $\pm$ std). GT uses ground-truth pose (upper bound).} 
\label{tab:manipulation}
\centering
\setlength{\tabcolsep}{3pt}
\footnotesize
\begin{tabular}{l|cccccc}
\toprule
\multirow{2}{*}{Pose Source} & \multicolumn{6}{c|}{Occlusion Ratio}\\ & 0\% & 10\% & 30\% & 50\% & 70\% & 90\% \\
\midrule
GT & 21.25 & --- & --- & --- & --- & --- \\
\midrule
V-only 
  & \small 0.46\tiny$\pm$.16 
  & \small 0.63\tiny$\pm$.06 
  & \small 0.83\tiny$\pm$.24 
  & \small 1.58\tiny$\pm$.34 
  & \small 1.97\tiny$\pm$1.0 
  & \small 1.52\tiny$\pm$.18 \\
Naive V+H 
  & \small 1.54\tiny$\pm$.47 
  & \small 1.40\tiny$\pm$.55 
  & \small 1.86\tiny$\pm$.58 
  & \small 1.53\tiny$\pm$.53 
  & \small 1.75\tiny$\pm$.40 
  & \small 2.00\tiny$\pm$.29 
    \\
FingerMLP 
  & \small 1.63\tiny$\pm$.06 
  & \small 0.61\tiny$\pm$.14 
  & \small 0.79\tiny$\pm$.67 
  & \small 1.48\tiny$\pm$.19 
  & \small 5.04\tiny$\pm$2.1 
  & \small 2.45\tiny$\pm$.38 
\\
16-token 
  & \small 0.27\tiny$\pm$.04 
  & \small 0.23\tiny$\pm$.03 
  & \small 0.33\tiny$\pm$.05 
  & \small 0.30\tiny$\pm$.04 
  & \small 0.29\tiny$\pm$.05 
  & \small 0.30\tiny$\pm$.10 
\\
\midrule
\textbf{KineFuse} 
  & \small \textbf{4.61}\tiny$\pm$1.3 
  & \small \textbf{4.47}\tiny$\pm$1.4 
  & \small \textbf{4.72}\tiny$\pm$.42 
  & \small \textbf{4.47}\tiny$\pm$.83 
  & \small \textbf{5.10}\tiny$\pm$.28 
  & \small \textbf{3.81}\tiny$\pm$.85 
\\
\bottomrule
\end{tabular}
\end{table}

\subsection{Pose Estimation across Occlusion Severity}\label{sec:exp_occ}

Our main experiment evaluates all models across five occlusion levels. Table~\ref{tab:manipulation} and Fig.~\ref{fig:qualitative} report the results.

\textbf{Per-frame refinement.} The results shows that all models achieve comparable per-frame accuracy (ADD $<$ 0.011\,cm), with a best-to-worst ratio of only 1.12 times. The visual backbone dominates single-step refinement, leaving little room for haptic encoders to differentiate. This motivates sequential evaluation.

\textbf{Sequential open-loop tracking.} When pose estimates are chained across frames (Table~\ref{tab:openloop}, Fig.~\ref{fig:qualitative}), architectural differences amplify dramatically. KineFuse achieves 4.04\,cm mean position error (2 times lower than V-only's 8.11\,cm) and maintains constant angular error at 47.6 degrees $\pm$ 0.1 degree regardless of occlusion level, while V-only fluctuates between 54.9 and 79.4 degrees. Note that the reorientation task spans a full SO(3) range (up to 180 degrees), making 47.6 degrees a meaningful reduction from the V-only baseline's 54.9–79.4 degrees range. At low occlusion ($0$--$20\%$), all models perform comparably, as the visual backbone provides sufficient pose information. As occlusion increases beyond $40\%$, vision-only tracking degrades sharply while haptic-augmented models maintain lower error. Among fusion models, our structured encoder consistently outperforms both the naive baseline and the 16-token variant, with the gap widening at higher occlusion. This confirms two points: (i)~sparse haptics become complementary to vision precisely when visual estimates degrade, and (ii)~structured encoding of these signals yields additional gains over naive concatenation.

Fig.~\ref{fig:qualitative} provides a qualitative comparison: KineFuse maintains accurate pose alignment through 35 tracking steps at 30\% occlusion, while the vision-only tracker diverges by step 20.
\begin{figure}[t]
\centering
\includegraphics[width=\linewidth]{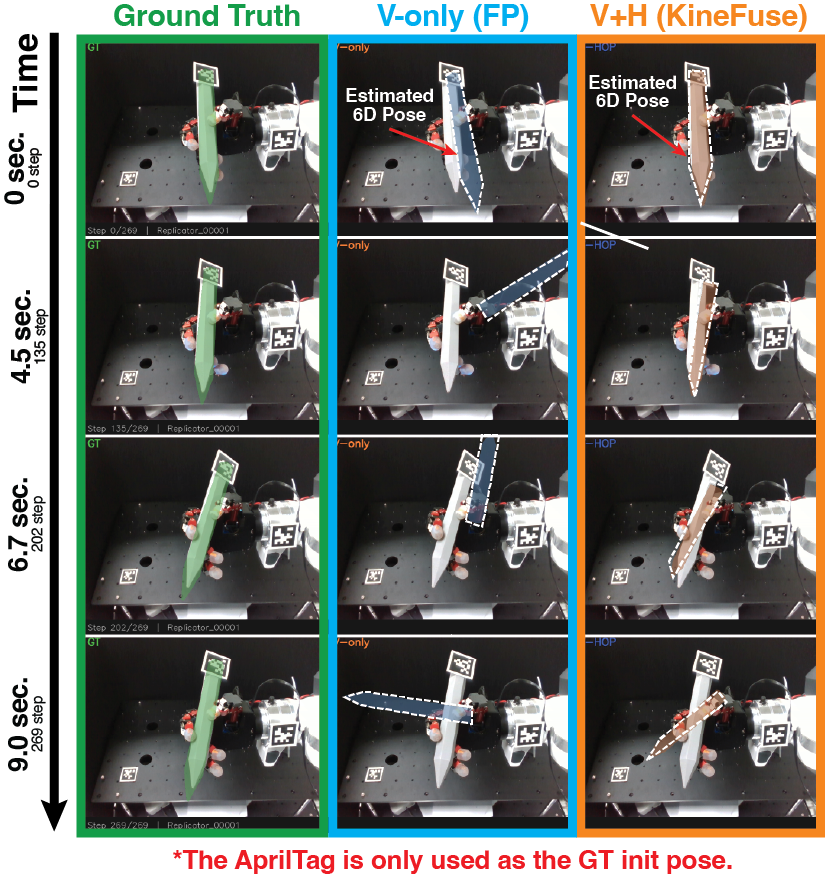}
\caption{Real-world qualitative results.}
\vspace{-0.5em}
\label{fig:real_demo}
\end{figure}

\begin{table}[t]
\centering
\caption{Pose tracking error.}
\label{tab:pose_error_rep01}
\begin{tabular}{c|cc|cc}
\toprule
\multirow{2}{*}{Step} & \multicolumn{2}{c|}{Trans.\ Error (mm)} & \multicolumn{2}{c}{Rot.\ Error (deg)} \\
& V-only & KineFuse & V-only & KineFuse \\
\midrule
0   &  11.9 & \textbf{1.4}  &   6.7 & \textbf{2.5}  \\
67  &  78.2 & \textbf{17.4} &  57.7 & \textbf{30.8} \\
135 & 108.2 & \textbf{35.9} & 130.3 & \textbf{63.5} \\
202 &  95.3 & \textbf{58.1} & 177.4 & \textbf{86.6} \\
269 &  96.6 & \textbf{85.4} &  \textbf{94.3} & 131.5 \\
\bottomrule
\end{tabular}
\end{table}
\vspace{-0.5em}

\subsection{Downstream Manipulation}\label{sec:exp_manip}

Accurate pose tracking is only valuable if it improves task performance. We evaluate this by deploying a separately trained RL policy for in-hand reorientation, replacing its ground-truth pose observation with estimated poses from our tracker. The policy is held fixed; only the pose source varies.


\textbf{Manipulation with estimated pose.} When deployed in a closed-loop manipulation pipeline, KineFuse tracking achieves higher success than V-only in both RL policies, consistent with its superior per-frame accuracy. However, cumulative tracking drift remains the primary bottleneck, reducing overall success to 9\% of the ground-truth upper bound (See Figure.~\ref{fig:tracking-err}). Addressing drift through temporal regularization or re-initialization strategies is left to future work.


By cross-referencing Fig.~\ref{fig:tolerrence_eval} and Table.~\ref{tab:manipulation}, we can trace manipulation failures directly to pose error magnitude: when V-only position error exceeds the tolerance threshold identified in Fig.~\ref{fig:tolerrence_eval}, task success drops correspondingly. Our model's lower pose error keeps it within the policy's operating range across a wider span of occlusion conditions.


\subsection{Ablation Studies}\label{sec:ablation}
\textbf{Structural vs. informational contribution.} A key question is whether haptic signals contribute through their runtime content or through the structural bias they impose during training. To disentangle these, we zero all haptic channels at inference. The result—nearly identical tracking to the full model, yet 2.3 times better than V-only provides direct evidence that the contribution is structural: the kinematic topology of the haptic encoder acts as an inductive bias that reshapes how the fusion transformer organizes visual attention, and this reorganization persists after haptic removal. This finding has practical implications: once trained with haptic data, the model can be deployed in scenarios where haptic sensors are temporarily unavailable (e.g., sensor failure, new end-effector) without performance degradation.

\subsection{Real-World Demonstration}\label{sec:real}

We demonstrate simple tracking performance on a physical setup using the same dexterous hand, a RealSense D435i camera, and a primitive tool object (see in Fig.~\ref{fig:concept}).

Since ground-truth poses are unavailable in the real-world setting, we report proxy metrics: temporal jitter (standard deviation of frame-to-frame position differences) and tracking failure count (frames where position jumps exceed). Fig.~\ref{fig:real_demo} shows representative frames. In addition, the results of Pose tracking error from AprilTag~\cite{olson2011apriltag} is in Table.~\ref{tab:pose_error_rep01}.








\section{Conclusion}
We studied how structuring sparse haptic signals affects vision-haptic fusion for in-hand pose tracking under occlusion. Three findings emerge: (i) per-frame evaluation cannot distinguish encoder quality, but sequential tracking amplifies differences by up to 15 times in downstream manipulation; (ii) the structured encoder learns task-specific gating—translation uses vision exclusively while rotation dedicates one attention head to haptic tokens—without explicit supervision; and (iii) this benefit is structural rather than informational, persisting even when haptic input is zeroed at inference.

\textbf{Limitations and future work.} Our evaluation is conducted on a single pencil-shaped object in simulation, with only qualitative real-world demonstration. We acknowledge this limits generalizability claims. However, we emphasize two architectural properties that support broader applicability: (i) the haptic encoder operates exclusively on hand-side signals (proprioception, F/T, contact) and encodes no object-specific features, making it object-agnostic by design; (ii) the URDF-aware spatial bias is derived from hand kinematics, not object geometry, and thus transfers to any object grasped by the same hand. Extending quantitative evaluation to diverse objects and real-world settings with ground-truth tracking remains our most important next step. Additionally, the modular perception-policy pipeline accumulates drift that the downstream policy cannot correct; joint end-to-end training could address this bottleneck.

\bibliographystyle{IEEEtran} 
\bibliography{ref/0_venues, ref/ref}
\end{document}